%% file: arxiv.tex
\documentclass{isprs} 
\usepackage{subfigure}
\usepackage{setspace}
\usepackage{geometry} 
\usepackage{epstopdf}
\usepackage[labelsep=period]{caption}  
\usepackage[british]{babel} 
\usepackage[hang]{footmisc}
\usepackage{booktabs}
\usepackage{multirow}
\usepackage{makecell}
\usepackage{url}
\usepackage[table]{xcolor}
\usepackage{siunitx}
\usepackage{array}
\usepackage{adjustbox}
\usepackage{amssymb}
\usepackage{graphicx}
\usepackage{amsmath}
\usepackage[dvipsnames]{xcolor}

\usepackage{pifont}       
\usepackage{bbding}       
\usepackage{fontawesome}  

\begin{document}

\title{ChangeDINO: DINOv3-Driven Building Change Detection in Optical Remote Sensing Imagery}
\date{}

\author{
 Ching Heng, Cheng\textsuperscript{1}, Chih Chung, Hsu\textsuperscript{2}}

\address{
	\textsuperscript{1 }Inst.\ of Data Science, National Cheng Kung University, \\Tainan, Taiwan - re6131016@gs.ncku.edu.tw\\
	\textsuperscript{2 }Inst.\ of Intelligent Systems College of Artificial Intelligence, \\National Yang Ming Chiao Tung University, Hsinchu, Taiwan - chihchung@nycu.edu.tw\\
}

\abstract{
Remote sensing change detection (RSCD) aims to identify surface changes from co-registered bi-temporal images. However, many deep learning–based RSCD methods rely solely on change-map annotations and underuse the semantic information in non-changing regions, which limits robustness under illumination variation, off-nadir views, and scarce labels. 
This article introduces \textbf{ChangeDINO}, an end-to-end multiscale Siamese framework for optical building change detection. The model fuses a lightweight backbone stream with features transferred from a frozen DINOv3, yielding semantic- and context-rich pyramids even on small datasets. A spatial–spectral differential transformer decoder then exploits multi-scale absolute differences as change priors to highlight true building changes and suppress irrelevant responses. Finally, a learnable morphology module refines the upsampled logits to recover clean boundaries. Experiments on four public benchmarks show that \textbf{ChangeDINO} consistently outperforms recent state-of-the-art methods in IoU and F1, and ablation studies confirm the effectiveness of each component. The source code is available at \url{https://github.com/chingheng0808/ChangeDINO}.
}


\maketitle
\sloppy

\input{sections/intro}

\input{sections/related}

\input{sections/methods}

\input{sections/experiments}

\input{sections/conclusion}

{
	\begin{spacing}{1.17}
		\normalsize
		\bibliography{ISPRSguidelines_authors} 
	\end{spacing}
}

\end{document}

%% file: sections/intro.tex
\section{Introduction}\label{INTRO}

Remote sensing change detection (RSCD) in multi-temporal remote sensing imagery is central to Earth observation. By comparing satellite or aerial images across time, RSCD reveals land-cover dynamics from natural and human activities ~\cite{landuse,cd_survey}. Buildings are an important focus area because changes inform urban planning, regulatory compliance, and risk assessment. With high-resolution aerial, drone , and satellite imagery, deep learning enables large-scale monitoring, detection of subtle structural changes, and data-driven support for infrastructure management and sustainable development ~\cite{optical_imagery_cd_survey}.

RSCD remains challenging due to cross-temporal and cross-domain variability. Image pairs may come from different sensors such as optical, multispectral, or SAR, or they may differ in illumination, seasonality, and viewing geometry even within one modality ~\cite{cdcd_survey}. These factors introduce spectral and geometric inconsistencies unrelated to true changes ~\cite{HUSSAIN201391}. Traditional hand-crafted methods struggle with such variability, while deep learning has become the prevailing approach by learning hierarchical, task-specific representations with stronger robustness and generalization ~\cite{fc-siam}.

However, current models face limited data scale and architectural constraints. Many RSCD datasets are small, geographically narrow, or task-specific labeled, encouraging overfitting and limiting access to global semantic context ~\cite{sample_survey}. Mainstream Siamese pipelines~\cite{fc-siam} fuse multiscale features with convolutional or transformer decoders, yet often miss fine pixel-level differences and are influenced by irrelevant context. Downsampling for efficiency and later upsampling also blurs boundaries.

To address these challenges, we propose \textbf{ChangeDINO}, a multiscale Siamese framework. It leverages the pretrained DINOv3 foundation model as the encoder, introduces a differential transformer–based decoder, originally proposed in the large-language-model domain~\cite{diff_trans}, to reason over cross-temporal context and suppress noise, and adopts a learnable morphological module for final mask refinement. The main contributions of ChangeDINO include:

\begin{itemize}
    \item Leverage DINOv3 pretrained in the encoder to inject semantically rich features without requiring task\mbox{-}specific semantic labels.
    
    \item Propose a differential transformer\mbox{–}based decoder that strengthens attention to relevant cross\mbox{-}temporal context for precise, pixel\mbox{-}level change modeling while suppressing distractors.
    
    \item Introduce a learnable, morphology\mbox{-}based refinement head with trainable structuring kernels that denoise predictions and sharpen subtle\mbox{-}change boundaries in end\mbox{-}to\mbox{-}end training.
\end{itemize}

%% file: sections/related.tex
\section{Related Works}\label{RELATED}

\subsection{Conventional Methods}

\begin{figure*}[htbp]
  \centering
  \includegraphics[width=\textwidth]{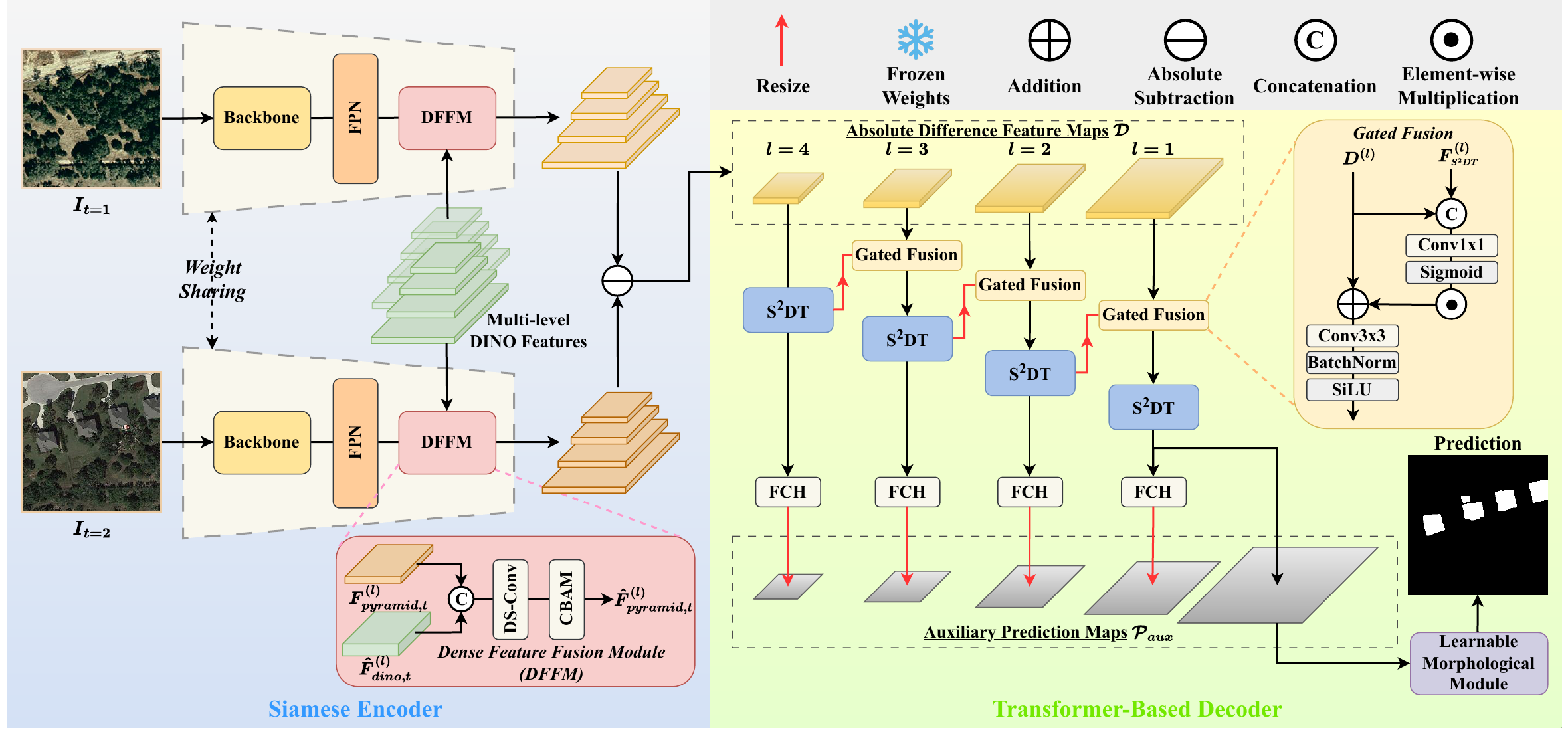}
    \caption{\textbf{Overall architecture of ChangeDINO.} The model adopts a classic multi-scale encoder–decoder and is trained end-to-end for optical building change detection. Please zoom-in for details.}
  \label{fig:arch-main}
\end{figure*}

Classical RSCD operates on bi-temporal imagery with pixel-wise algebra or statistics~\cite{cdcd_survey,cd_survey}, including image differencing and ratio-based transforms that highlight radiometric shifts but require careful thresholds and are sensitive to illumination and registration~\cite{diif_cd,cd_tech,ratio-cd}. To move beyond raw pixels, feature transformations nonlinearly project data to better separate changes~\cite{pixel_trans,graph_based}. Representative techniques include Change Vector Analysis (CVA), which aggregates multi-band distances~\cite{cva}, and PCA-like transforms that emphasize change-related variance but are scene-dependent and often yield binary, non-semantic maps~\cite{pca}. Alternatively, post-classification comparison (PCC) classifies each date and then compares the maps to produce a class-to-class change matrix, at the risk of propagating classification errors~\cite{pcc}. To improve spatial consistency, OBIA segments images into geo-objects prior to comparison~\cite{geo-obj}, and morphological operators are widely used as post-processing to denoise and regularize boundaries~\cite{cd_morph}.

\subsection{Deep Learning-Based Methods}

Deep learning reframed RSCD as a supervised segmentation task on image pairs. Early CNNs introduced end-to-end pipelines: FC-EF concatenates bi-temporal inputs for early fusion, whereas FC-Siam preserves two streams and fuses features later~\cite{fc-siam}. Building on this paradigm, subsequent Siamese designs improved multi-scale alignment and locality, for example SNUNet-CD~\cite{snunet} with nested dense skips and IFNet~\cite{IFNet} with enriched fusion. Broadly, two families now dominate: \emph{difference-based} methods that inject explicit image or feature differences to guide attention toward changes~\cite{a2net,clafa}, and \emph{fusion-based} methods that concatenate or attend across scales and times to learn discriminative joint representations~\cite{cgnet,bifa}.

To better capture long-range dependencies, transformer architectures further advance global context modeling. BIT~\cite{bit} couples CNN features with a transformer encoder over bi-temporal tokens, and ChangeFormer~\cite{changeformer} employs hierarchical vision transformers with lightweight decoders, improving cross-temporal interaction over CNN baselines. Recent works also explore the role of data scale and priors: foundation-model approaches such as ChangeCLIP~\cite{changeclip} adapt vision-language pretraining to emphasize semantic relevance and reduce sensitivity to seasonal or illumination shifts. In parallel, state-space models (SSMs) offer linear-time global modeling, with Mamba-based RSCD variants reporting transformer-level accuracy and improved efficiency via selective scanning~\cite{cdmamba,changemamba}. Furthermore, self-supervised methods~\cite{adv_cd,adv_cd2,cond_gan} employ domain-adaptation to reduce shifts across sensors and seasons, while GAN-based augmentation~\cite{aug_gan} synthesizes realistic change samples to improve overall RSCD performance.

Overall, the literature has progressed from thresholded pixel algebra to context-aware neural architectures that fuse multi-scale features and model long-range relations. Yet open issues persist in cross-domain generalization, fine-grained boundary fidelity, and data efficiency~\cite{optical_imagery_cd_survey,sample_survey}, motivating methods that combine strong priors from large-scale pretraining with architectures tailored for precise differential reasoning and morphology-aware refinement.

%% file: sections/methods.tex
\section{Methodology}\label{METHOD}

\subsection{Method Overview}
As illustrated in Fig.~\ref{fig:arch-main}, our pipeline takes a pair of cross-temporal optical images and processes them with a Siamese encoder, which combines the pretrained DINOv3 and a lightweight backbone with a Feature Pyramid Network (FPN). The encoder yields a multi-scale feature pyramid that is semantically rich and relatively domain-agnostic, emphasizing building structures while remaining robust to illumination and seasonal variation.

From the two pyramids, we construct a multi-scale change prior by taking absolute differences at each resolution as change priors. This  multi-scale prior features are fed into a cascade-style Differential Transformer–based decoder that combines spatial and spectral-wise (channel) self-attention. The decoder focuses on truly changed regions and suppresses distractors. At each scale, fully convolution heads produce auxiliary prediction maps to stabilize optimization and guide progressive refinement.

Finally, a learnable, morphology-based refinement head performs shape-preserving refinement on the last auxiliary prediction logit, improving boundary sharpness and object connectivity to produce final prediction. The entire network is trained end-to-end for optical building change detection. The details of each component are described in the following subsections.

\begin{figure}[htbp]
  \centering
  \includegraphics[width=0.45\textwidth]{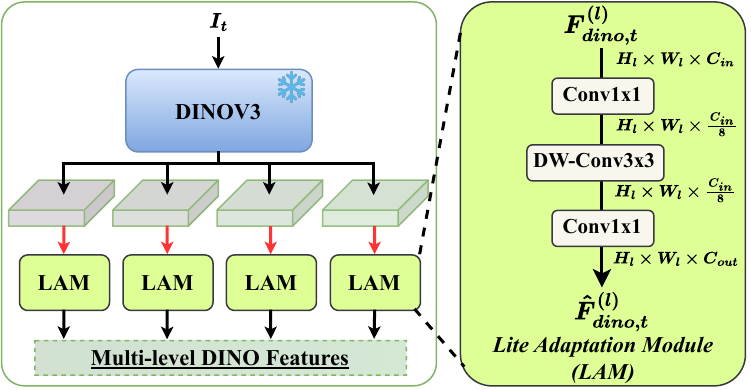}
  \caption{Lightweight feature adapter aligning DINOv3 features with our backbone.}
  \label{fig:arch-dino}
\end{figure}

\subsection{Semantic and Context-Rich Multi-Scale Features via Siamese Encoder with DINOv3 Pretraining}

Given a cross-temporal optical pair $\{I_t\}_{t\in\{1,2\}}$, we adopt a lightweight convolutional backbone (MobileNet~\cite{mobilenet}) followed by an FPN~\cite{clafa} to construct multi-scale representations. Let the backbone and FPN be $\Phi_{\mathrm{backbone}}$ and $\Phi_{\mathrm{fpn}}$, respectively. For each time $t$, the backbone extracts per-level features that are aggregated by the FPN into a top–down pyramid:
\begin{equation}
\mathcal{F}_{\mathrm{pyramid},t}=\big\{F^{(l)}_{\mathrm{pyramid},t}\big\}_{l=1}^{4}
=\Phi_{\mathrm{fpn}}\!\big(\Phi_{\mathrm{backbone}}(I_t)\big),
\label{eq:pyramid}
\end{equation}
where $l$ indexes the pyramid levels.

When trained on limited data without strong semantic supervision, the pyramid features $\mathcal{F}_{\mathrm{pyramid},t}$ tend to lack context. To inject semantic priors, we incorporate the large-scale pretrained foundation model DINOv3~\cite{dinov3}. To avoid catastrophic forgetting, DINOv3 is frozen (Fig.~\ref{fig:arch-dino}) and four intermediate, semantics-rich feature maps are tapped, resized to the corresponding pyramid scales, and passed through a \emph{Lite Adaptation Module} (LAM) to align channels and distill semantics:
\begin{equation} 
    {\mathcal{F}}_{\mathrm{dino},t} = \mathrm{DINOv3}(I_t) = \{\,F^{(l)}_{\mathrm{dino},t}\,\}_{l=1}^{4} = {\mathcal{F}}_{\mathrm{dino},t}, 
    \label{eq:dino1} 
\end{equation} 
\begin{equation} 
    \hat{\mathcal{F}}_{\mathrm{dino},t}=\{\,\hat{F}^{(l)}_{\mathrm{dino},t}\,\}_{l=1}^{4}=\{\,\Phi^{(l)}_{\mathrm{LAM}}(\mathrm{Resize}(\,F^{(l)}_{\mathrm{dino},t}\,))\}_{l=1}^{4}, 
    \label{eq:dino2} 
\end{equation} 
\noindent
where $\mathrm{DINOv3}(\cdot)$ denotes the frozen DINOv3 model, $\Phi^{(l)}_{\mathrm{LAM}}(\cdot)$ the $l$-th lightweight adapter, and $\mathrm{Resize}(\cdot)$ denotes bilinear interpolation resizing.

To fuse the adapted DINO features with the backbone–FPN pyramid, we employ a \emph{Dense Feature Fusion Module} (DFFM) that concatenates the two streams, applies a depthwise–separable convolution~\cite{xception}, and uses CBAM attention~\cite{cbam} to produce a semantic- and context-rich pyramid:
\begin{equation}
\hat{\mathcal{F}}_{\mathrm{pyramid},t}^{(l)}
=\Phi^{(l)}_{\mathrm{DFFM}}\!\big(F^{(l)}_{\mathrm{pyramid},t},\,\hat{F}^{(l)}_{\mathrm{dino},t}\big),
\;\; l=1,\dots,4,
\end{equation}
where $F^{(l)}_{\mathrm{pyramid},t}$ and $\hat{F}^{(l)}_{\mathrm{dino},t}$ are defined in Eqs.~\eqref{eq:pyramid} and~\eqref{eq:dino2}.

At the end of the encoder, we compute multi-scale element-wise absolute differences $\mathcal{D}=\{D^{(l)}\}_{l=1}^{4}$ between the bi-temporal pyramids to obtain the decoder inputs:
\begin{equation}
    D^{(l)}=\left\lvert \hat{\mathcal{F}}_{\mathrm{pyramid},1}^{(l)}-\hat{\mathcal{F}}_{\mathrm{pyramid},2}^{(l)} \right\rvert.
    \label{eq:absdiff}
\end{equation}

Overall during training, the backbone+FPN branch adapts to task-specific details and local structures, while the frozen DINOv3 branch supplies robust semantic context; their fusion yields multi-scale representations that are both fine-grained and semantically consistent, from which we compute per-level absolute differences as a multi-scale change prior to drive the decoder.

\begin{figure}[htbp]
  \centering
  \includegraphics[width=0.35\textwidth]{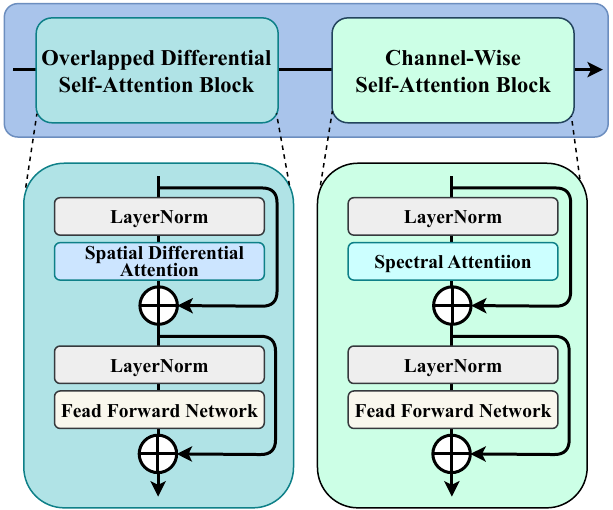}
  \caption{\textbf{Spatial–spectral differential transformer ($\mathrm{S}^2\mathrm{DT}$) block.} Incorporates a differential transformer into overlapped-window spatial self-attention and pairs it with channel-wise self-attention to refine feature intensities.}
  \label{fig:arch-ssdt}
\end{figure}

\subsection{Building-Aware Decoder via Spatial and Spectral Self-Attention Decoder}

The decoder upsamples and transforms the fused bi\mbox{-}temporal features, specifically the multi\mbox{-}scale absolute\mbox{-}difference maps $\mathcal{D}$ (Eq.~\ref{eq:absdiff}), into a high\mbox{-}resolution change map. It reconstructs spatial detail and produces a pixel\mbox{-}wise prediction that separates changed from unchanged regions.

Guided by $\mathcal{D}$ as a change prior, we design a Spatial–Spectral Differential Transformer ($\mathrm{S}^2\mathrm{DT}$), a transformer module that fuses \emph{spatial} and \emph{spectral (channel\mbox{-}wise)} self\mbox{-}attention and instantiates the differential transformer~\cite{diff_trans} as the core attention mechanism. Differential transformers, originally validated in large\mbox{-}language\mbox{-}model settings for focusing on informative tokens while suppressing noise, are well suited here for filtering illumination\mbox{-}induced or misregistration artifacts and other irrelevant responses.

The spatial differential attention in the differential self\mbox{-}attention block of $\mathrm{S}^2\mathrm{DT}$ is summarized as follows. Given a feature map $X\in\mathbb{R}^{C\times H\times W}$, we obtain queries, keys, and values via $1{\times}1$ convolutions and reshape them into $h$ heads with token length $N{=}HW$ and head width $d{=}C/h$:
\begin{equation}
\begin{aligned}
&Q,K,V \in \mathbb{R}^{h\times N\times 2d},\\
&Q=[Q_1;\,Q_2],\quad K=[K_1;\,K_2],
\end{aligned}
\end{equation}
where $[\cdot;\cdot]$ denotes channel splitting. Two spatial attentions are computed on the halves:
\begin{equation}
A_1=\mathrm{softmax}\!\left(\frac{Q_1K_1^\top}{\sqrt{d}}\right),\quad
A_2=\mathrm{softmax}\!\left(\frac{Q_2K_2^\top}{\sqrt{d}}\right).
\end{equation}
They are combined in a multihead differential form:
\begin{equation}
\begin{aligned}
\widetilde{X} &= \bigl(A_1-\Lambda A_2\bigr)\,V \in \mathbb{R}^{h\times N\times 2d},\\
\widetilde{X} &= \bigl(\widetilde{X}^{(i)}\bigr)_{i=1}^{h},\qquad
\widetilde{X}^{(i)} \in \mathbb{R}^{N\times 2d},
\end{aligned}
\end{equation}
where $\Lambda=\mathrm{diag}(\lambda^{(1)},\dots,\lambda^{(h)})$ holds per\mbox{-}head, positive, learnable coefficients. The per\mbox{-}head outputs are normalized with RMSNorm~\cite{rmsnorm}, concatenated, and projected back to the spatial tensor. The spatial differential attention operator $\Phi_{\mathrm{SDA}}$ is defined as:
\begin{equation}
\Phi_{\mathrm{SDA}}(X)=\mathbf{W}_{\mathrm{prj}}\,
\mathrm{Concat}_{i=1}^h\!\bigl(\mathrm{RMSNorm}(\widetilde{X}^{(i)})\bigr)\in\mathbb{R}^{C\times H\times W},
\end{equation}
where $\mathbf{W}_{\mathrm{prj}}$ denotes the projection weights. This differential attention $(A_1-\Lambda A_2)$ emphasizes informative spatial correspondences while attenuating distractors, yielding sharper and cleaner responses for change localization.

Consequently, $\mathrm{S}^2\mathrm{DT}$ targets pixel\mbox{-}level change discrimination across pyramid levels. To coordinate information across scales, we adopt a gated\mbox{-}fusion operator $G(\cdot)$, illustrated in Fig.~\ref{fig:arch-main}, that adaptively controls cross\mbox{-}level contributions. For level $l$,
\begin{equation}
F^{(l)}_{\mathrm{S^2DT}} =
\begin{cases}
G\!\big(D^{(l)},\,F^{(l+1)}_{\mathrm{S^2DT}}\big), & l=1,2,3,\\[2pt]
\Phi_{\mathrm{S^2DT}}^{(l)}\!\big(D^{(l)}\big), & l=4,
\end{cases}
\label{eq:s2dt}
\end{equation}
where $\Phi_{\mathrm{S^2DT}}^{(l)}(\cdot)$ denotes the $\mathrm{S}^2\mathrm{DT}$ block at level $l$. Fully convolutional heads (FCHs) then produce auxiliary predictions for deep supervision:
\begin{equation}
\hat{P}_{\mathrm{aux}}^{(l)} = \mathrm{Upsample}\!\left(H^{(l)}\!\left(F^{(l)}_{\mathrm{S^2DT}}\right)\right), \qquad l=1,\dots,4,
\end{equation}
where $H^{(l)}(\cdot)$ denotes the oer-level FCH that maps $\mathrm{S}^2\mathrm{DT}$ features to a binary change logit map, and $\mathrm{Upsample}(\cdot)$ denotes bilinear interpolation.


\subsection{Refining Binary Prediction Using Learnable Morphology}

Direct upsampling via interpolation can produce a reasonably structured change logit, but it may still contain fragmented interiors and spurious protrusions. Classical morphology helps alleviate this issue, but fixed structuring elements often over-smooth edges or remove fine details. We therefore propose learnable morphological module (LMM), which adopt \emph{learnable} structuring elements (fixed window size, trainable weights) and fuse the refined result with the original logits in an end-to-end manner. The final prediction $\hat{P}$ is
\begin{equation}
\begin{split}
\hat{P}
&= \alpha\,\sigma^{-1}\!\left(\mathrm{Closing}\!\left(\mathrm{Opening}\!\left(\sigma(\hat{P}_{\mathrm{aux}}^{(1)}),\Omega_1\right),\Omega_2\right)\right) \\
&\quad +(1-\alpha)\,\hat{P}_{\mathrm{aux}}^{(1)} .
\end{split}
\end{equation}
where $\alpha\in[0,1]$ is a learnable mixing weight; $\sigma(\cdot)$ denotes the sigmoid function and $\sigma^{-1}(\cdot)$ its inverse (logit), and $\Omega_1,\Omega_2$ are learnable structuring kernels with window sizes of 3 and 5, respectively. We define opening and closing as
\begin{equation}
\mathrm{Opening}(M,\Omega)=\mathrm{Dil}\!\big(\mathrm{Ero}(M,\Omega),\,\Omega\big),
\end{equation}
\begin{equation}
\mathrm{Closing}(M,\Omega)=\mathrm{Ero}\!\big(\mathrm{Dil}(M,\Omega),\,\Omega\big),
\end{equation}
with $\mathrm{Ero}(\cdot,\cdot)$ and $\mathrm{Dil}(\cdot,\cdot)$ denoting differentiable erosion and dilation implemented via softmin/softmax approximations. This morphology head removes noise while preserving building shapes and connectivity.

Finally, we obtain the binary change map $\hat{P}_{\mathrm{bin}}$ by thresholding the logit prediction $\hat{P}$.

\begin{figure}[htbp]
  \centering
  \includegraphics[width=0.26\textwidth]{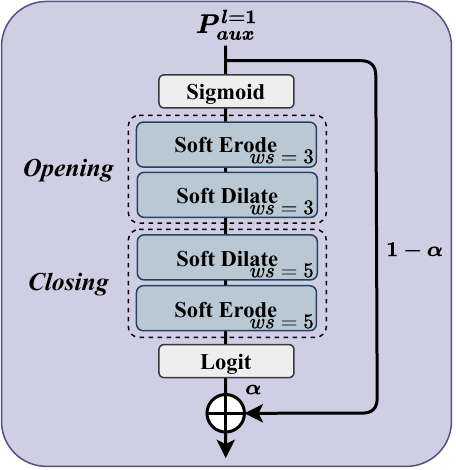}
  \caption{\textbf{Learnable morphological module (LMM).} Classical opening and closing with learnable structuring elements further refine the prediction.}
  \label{fig:arch-morph}
\end{figure}

\subsection{Objective Function}

Let $Y\in\{0,1\}^{H\times W}$ denote the binary ground truth, $\hat{P}\in\mathbb{R}^{H\times W}$ the final prediction logits, and $\{\hat{P}_{\mathrm{aux}}^{(l)}\}_{l=0}^{4}$ the auxiliary prediction logits. To address the foreground–background class imbalance, we use two popular losses, Focal loss~\cite{focal} $\mathcal{L}_{\mathrm{focal}}$ and Dice loss~\cite{dice} $\mathcal{L}_{\mathrm{dice}}$, and weight them with a hyperparameter $\beta$. We also apply deep supervision to the auxiliary maps:
\begin{equation}
\mathcal{L}_{\mathrm{main}}
=\mathcal{L}_{\mathrm{focal}}(\hat{P},Y)
+\beta\,
\mathcal{L}_{\mathrm{dice}}(\hat{P},Y),
\end{equation}
\begin{equation}
\mathcal{L}_{\mathrm{aux}}
=\sum_{l=0}^{4} \Big[
\mathcal{L}_{\mathrm{focal}}(\hat{P}_{\mathrm{aux}}^{(l)},Y)
+\beta\,
\mathcal{L}_{\mathrm{dice}}(\hat{P}_{\mathrm{aux}}^{(l)},Y)
\Big],
\end{equation}
\begin{equation}
\mathcal{L}_{\mathrm{total}}
=\mathcal{L}_{\mathrm{main}}
+0.5\,\mathcal{L}_{\mathrm{aux}}.
\end{equation}
We set $\beta=0.5$ for all experiments.

%% file: sections/experiments.tex
\definecolor{cyan}{rgb}{0.0, 1.0, 1.0}

\section{Experiments}\label{EXP}

\subsection{Experiment Details}
\paragraph{Benchmark Datasets.}
We evaluate on four public change detection datasets: LEVIR\mbox{-}CD~\cite{levir}, WHU\mbox{-}CD~\cite{whu}, S2Looking\mbox{-}CD~\cite{s2looking}, and SYSU\mbox{-}CD~\cite{sysu}, the cross-temporal image pairs in all datasets are well registered. Using the same data and official splits as in previous work~\cite{cgnet} to ensure a fair comparison of apples to apples. LEVIR\mbox{-}CD, WHU\mbox{-}CD, and S2Looking\mbox{-}CD are object-specific, focusing on building change detection (BCD), whereas SYSU\mbox{-}CD is category-agnostic. LEVIR\mbox{-}CD contains Google Earth image-pair patches collected across 20 regions with 5–14 year intervals. WHU\mbox{-}CD comprises aerial images of the same area before and after an earthquake. S2Looking\mbox{-}CD consists of large side-looking satellite pairs captured at different off-nadir angles, with significant illumination variation and extensive rural scenes. SYSU\mbox{-}CD includes diverse change types such as road expansion, newly built urban structures, vegetation changes, suburban sprawl, and groundwork prior to construction. The numbers of training/validation/testing pairs are 7{,}120/1{,}024/2{,}048 (LEVIR\mbox{-}CD), 4{,}536/504/2{,}760 (WHU\mbox{-}CD), 56{,}000/8{,}000/16{,}000 (S2Looking\mbox{-}CD), and 12{,}000/4{,}000/4{,}000 (SYSU\mbox{-}CD). Following common practice, we tile image pairs into $256\times256$ patches without overlap and apply no additional curation beyond the official partitions.

\begin{figure*}[htbp]
  \centering
  \includegraphics[width=\textwidth]{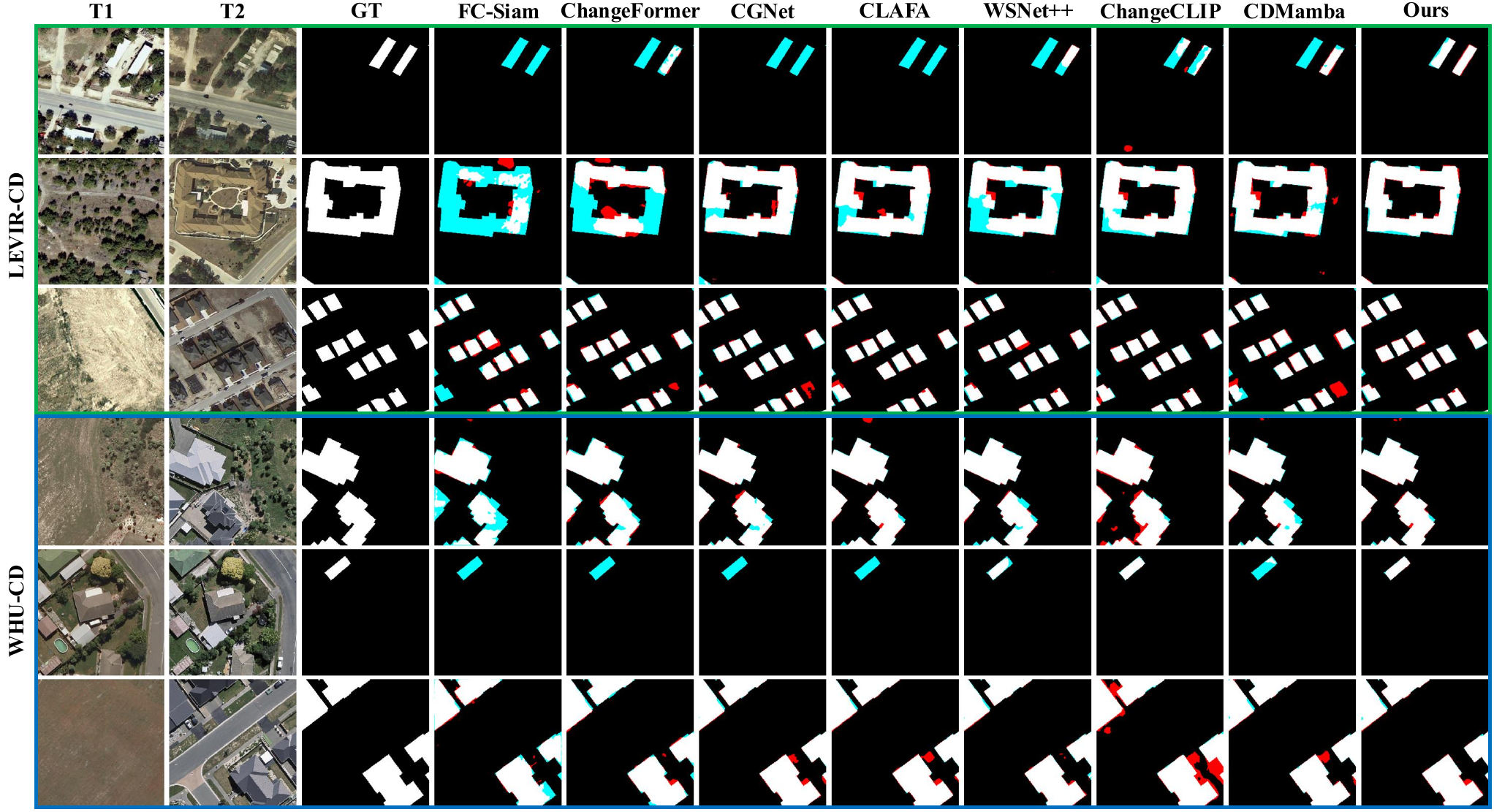}
    \caption{Qualitative experimental results on LEVIR\mbox{-}CD and WHU\mbox{-}CD.}
  \label{fig:levir-whu}
\end{figure*}

\input{tables/exp1}

\input{tables/exp2}

\paragraph{Evaluation Metrics.}
For the binary change detection performance evaluation, we use the Intersection-over-Union (IoU), F1 score, precision (Pre.) and recall (Rec.) for the change class.

\paragraph{Compared Methods.} 
To evaluate the proposed \textbf{ChangeDINO}, we compare it against a broad set of state-of-the-art RSCD methods. As noted in \emph{Related Work} section, recent RSCD research is dominated by deep learning, primarily along two lines: CNN-based and Transformer-based models. Accordingly, we benchmark against multiple representative approaches: five CNN-based (FC-Diff~\cite{fc-siam}, IFNet~\cite{IFNet}, A2Net~\cite{a2net}, CGNet~\cite{cgnet}, CLAFA~\cite{clafa}) and four Transformer-based (BIT~\cite{bit}, ChangeFormer~\cite{changeformer}, BiFA~\cite{bifa}, ChangeRD~\cite{changerd}). In addition, we include three recent frameworks that reflect emerging trends: a multi-model-based method (ChangeCLIP~\cite{changeclip}), a domain-adaptation method (WS-Net++~\cite{wsnet}), and a Mamba-based method (CDMamba~\cite{cdmamba}).

\paragraph{Implementation Details.} The proposed ChangeDINO is implemented in PyTorch~\cite{pytorch} and trained/evaluated on a single NVIDIA RTX 3090 GPU. During training, input pairs are cropped to $256\times256$ with a batch size of 16. We use the AdamW optimizer~\cite{adamw} with an initial learning rate of $5\times10^{-4}$ (set to $1\times10^{-4}$ for WHU\mbox{-}CD), and apply a cosine decay schedule down to $1\times10^{-7}$. Models are trained for 100 epochs on LEVIR\mbox{-}CD and WHU\mbox{-}CD, and for 50 epochs on SYSU\mbox{-}CD and S2Looking\mbox{-}CD. Standard data augmentation (random rotations, flips, and crops) is applied during training.

\begin{figure*}[htbp]
  \centering
  \includegraphics[width=\textwidth]{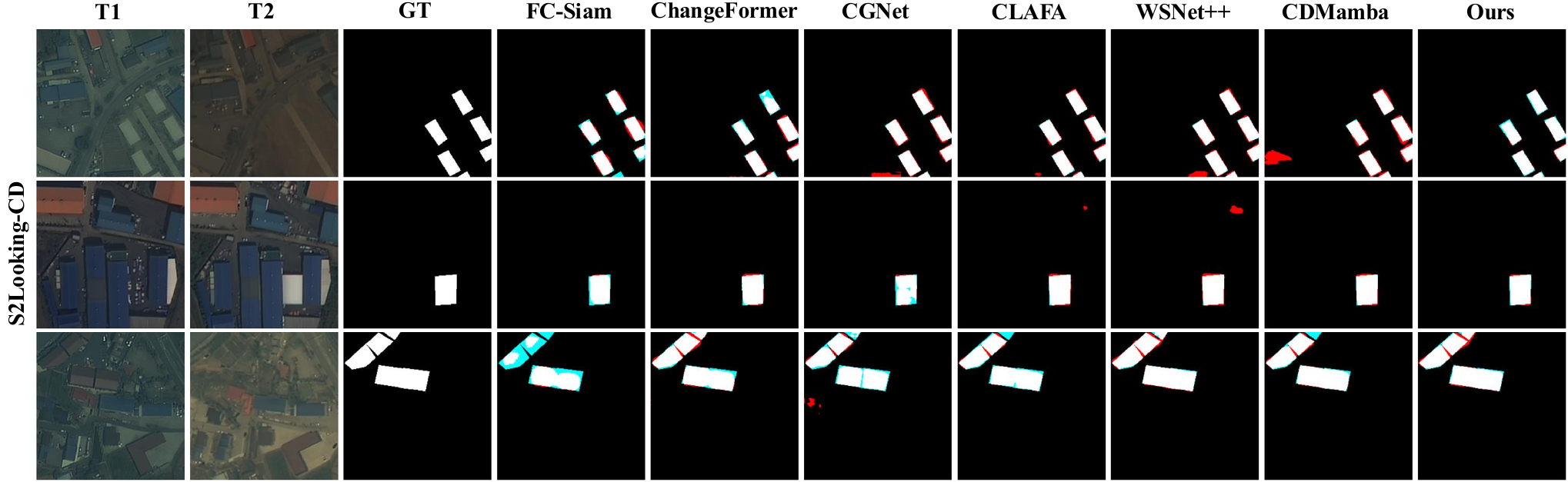}
    \caption{Qualitative experimental results on S2Looking\mbox{-}CD.}
  \label{fig:s2looking}
\end{figure*}
\begin{figure*}[htbp]
  \centering
  \includegraphics[width=\textwidth]{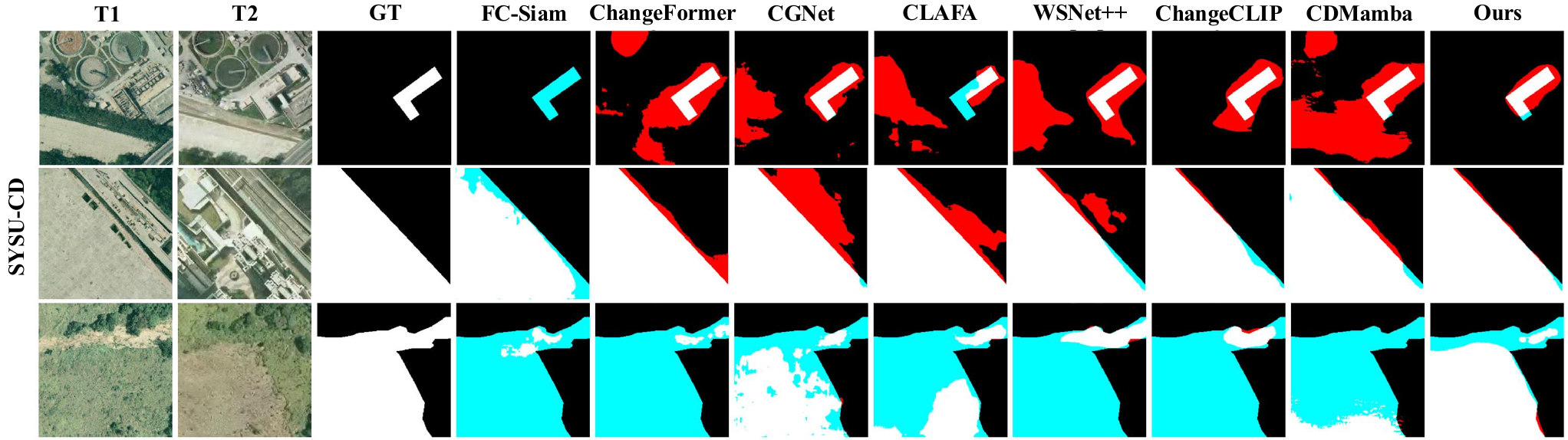}
    \caption{Qualitative experimental results on SYSU\mbox{-}CD}
  \label{fig:sysu}
\end{figure*}

\subsection{Quantify Analysis and Visualize Results}

In quantitative comparisons, the \textbf{best} and \underline{second best} results are highlighted in bold and underline, respectively, for clarity. LEVIR\mbox{-}CD and WHU\mbox{-}CD are comparatively easier benchmarks, whereas S2Looking\mbox{-}CD and SYSU\mbox{-}CD are more challenging: the former involves large side\mbox{-}looking imagery with varying off\mbox{-}nadir angles, and the latter is category\mbox{-}agnostic with diverse change types. Tables~\ref{tab:main1} and~\ref{tab:main2} report the quantitative results of our method against recent state of the art. An asterisk ($^*$) on ChangeCLIP indicates the use of the officially released pretrained weights only. Across all four datasets, \textbf{ChangeDINO} outperforms all SOTAs and attains the best IoU and F1 scores. CLAFA performs strongly on LEVIR\mbox{-}CD and SYSU\mbox{-}CD but lags on S2Looking\mbox{-}CD, while WS\mbox{-}Net++ excels on WHU\mbox{-}CD and S2Looking\mbox{-}CD yet trails on LEVIR\mbox{-}CD. ChangeCLIP is also highly competitive, reinforcing that foundation/model\mbox{-}pretrained representations are an effective and growing direction for RSCD. It is worth noting that FC\mbox{-}Diff and IFNet achieve high precision (Pre.) but underperform on other metrics, likely because they favor only the most salient changes and miss subtle or boundary pixels.

For qualitative comparison, due to space constraints, we select representative methods spanning different RSCD paradigms to visualize predictions. As shown in Fig.~\ref{fig:levir-whu}, on LEVIR\mbox{-}CD and WHU\mbox{-}CD our method exhibits the most accurate delineation of changes with fewer false positives (FP) and false negatives (FN). Note that true positives (TP) and true negatives (TN) are rendered in white and black, respectively, while FP and FN are rendered in \colorbox{Red}{\ } and \colorbox{cyan}{\ }, respectively. Figs.~\ref{fig:s2looking} and~\ref{fig:sysu} present results on the more challenging benchmarks. ChangeDINO maintains superior visual quality, producing cleaner masks with sharper building boundaries and less spurious noise.

\begin{figure}[htbp]
  \centering
  \includegraphics[width=0.48\textwidth]{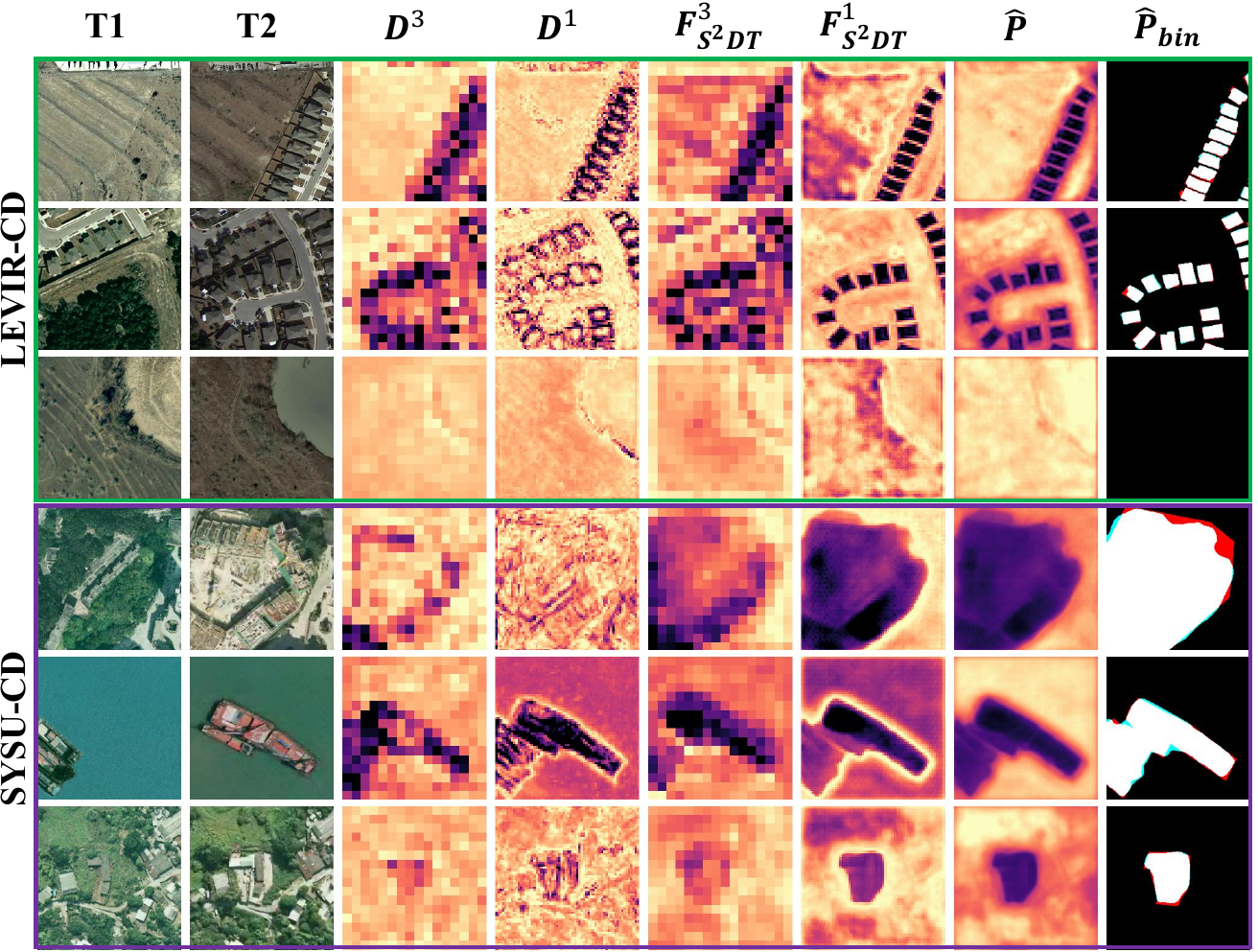}
    \caption{Visualized features of scenes from LEVIR\mbox{-}CD and SYSU\mbox{-}CD. Darker colors indicate stronger attention.  (Zoom-in for details).}
  \label{fig:feats}
\end{figure}

To further illustrate the effect of each component, we select six representative scenes from LEVIR\mbox{-}CD and SYSU\mbox{-}CD and visualize (i) the difference features at two pyramid levels ($D^{3}$ and $D^{1}$), (ii) the corresponding $\mathrm{S}^2\mathrm{DT}$ features, and (iii) the final prediction after LMM ($\hat{P}$), as shown in Fig.~\ref{fig:feats}. PCA is used for all visualizations. The difference maps show that deep levels already highlight potential change regions, while shallow levels better preserve building structure and suppress irrelevant objects. After the $\mathrm{S}^2\mathrm{DT}$ blocks, true changes become more distinct and noise is reduced. The LMM output further sharpens boundaries and produces cleaner masks.

\begin{figure}[htbp]
  \centering
  \includegraphics[width=0.485\textwidth]{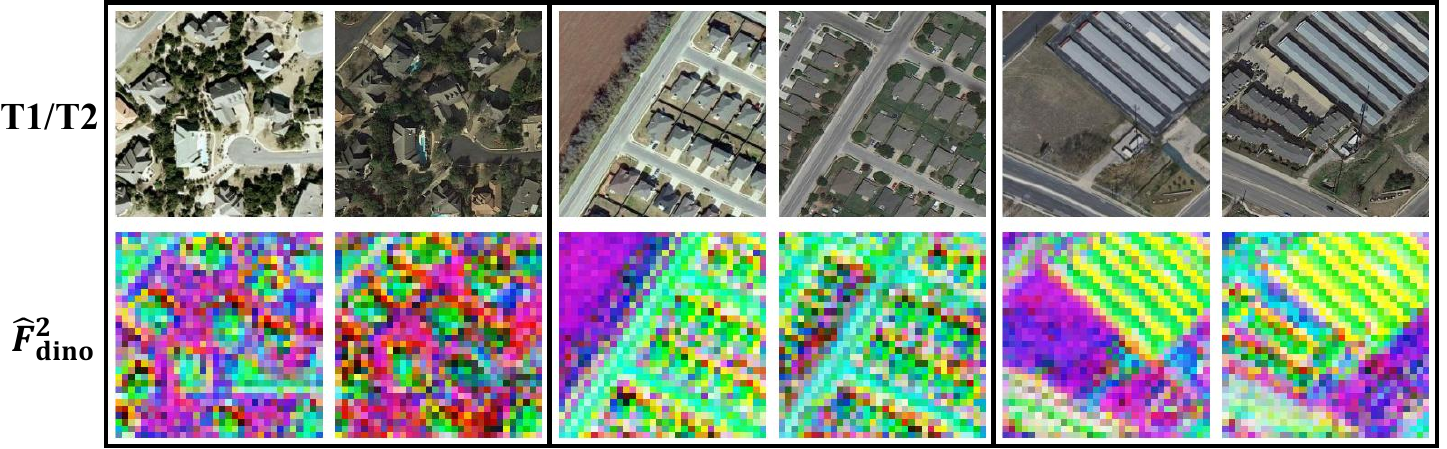}
    \caption{Visualized DFFM features (adapted DINOv3 features) from LEVIR\mbox{-}CD. We utilize vibrant color to demonstration. (Zoom-in for details).}
  \label{fig:feats_dino}
\end{figure}

\input{tables/ablation}

In addition, Fig.~\ref{fig:feats_dino} visualizes level\mbox{-}2 DFFM features of three image pairs (also via PCA). Different land covers such as trees, bare land, and roads are clearly separated, indicating that the DINOv3 branch supplies rich and discriminative semantics beyond the target buildings.

\subsection{Ablation Study}

To evaluate the contribution of each component in the proposed architecture, we conduct ablation experiments on the LEVIR\mbox{-}CD and SYSU\mbox{-}CD datasets. As shown in Tab.~\ref{tab:abl}, removing the DFFM causes the performance to drop by about 1.23 and 1.85 points of IoU on LEVIR\mbox{-}CD and SYSU\mbox{-}CD, respectively. This suggests that DFFM is the most influential of the three modules, and confirms that distilling semantics from a large\mbox{-}scale pretrained model is particularly beneficial when training on relatively small RSCD datasets. For the $\mathrm{S}^2\mathrm{DT}$ decoder, the “w/o $\mathrm{S}^2\mathrm{DT}$” variant is implemented by replacing it with a residual convolutional block. The observed degradation further indicates the effectiveness of combining spatial and spectral attention with the differential transformer design.

For the LMM, we find that enabling it improves results on both LEVIR\mbox{-}CD and SYSU\mbox{-}CD. It helps suppress small spurious responses while preserving building shapes, which is useful for both finely annotated data (LEVIR\mbox{-}CD) and coarser, large\mbox{-}area changes (SYSU\mbox{-}CD). Overall, the morphology\mbox{-}based refinement acts as an effective post-prediction regularizer across different annotation granularities. Overall, these ablations demonstrate that each proposed component in \textbf{ChangeDINO} contributes to improving RSCD performance under different dataset characteristics.

%% file: tables/exp1.tex

\begin{table}[ht]
\centering
\setlength{\tabcolsep}{3.5pt}
\caption{Quantitative comparisons in terms of IoU, F1, Recall, and Precision on LEVIR-CD and WHU-CD datasets.}
\label{tab:main1}
\begin{adjustbox}{width=0.48\textwidth}
\renewcommand{\arraystretch}{1.25}
\begin{tabular}{l|cccc|cccc}
\toprule
 & \multicolumn{4}{c|}{\textbf{LEVIR-CD}} & \multicolumn{4}{c}{\textbf{WHU-CD}} \\
\midrule
\midrule
Methods & IoU & F1 & Pre. & Rec. & IoU & F1 & Pre. & Rec. \\
\midrule
\midrule

FC-Diff & 73.23 & 84.55 & 89.18 & 80.37 & 63.63 & 77.78 & 89.29 & 68.89 \\
IFNet & 84.51 & 91.60 & \textbf{93.63} & 89.66 & 81.52 & 89.82 & 87.47 & \underline{92.30} \\
BIT & 81.72 & 89.94 & 90.33 & 89.56 & 68.02 & 80.97 & 74.01 & 89.37 \\
ChangeFormer & 81.69 & 89.92 & 91.70 & 88.20 & 78.51 & 87.96 & 91.07 & 85.06 \\
A2Net & 84.30 & 91.48 & 92.10 & 90.87 & 86.66 & 92.85 & 95.19 & 90.62 \\
CGNet & 85.21 & 92.01 & 93.15 & 90.90 & 86.21 & 92.59 & 94.47 & 90.79  \\
CLAFA & \underline{85.31} & \underline{92.07} & \underline{93.26} & 90.91 & \underline{87.80} & \underline{93.50} & \underline{95.51} & 91.58 \\
BiFA & 82.65 & 90.50 & 91.56 & 89.46 & 86.75 & 92.91 & 94.41 & 91.45 \\
WS-Net++ & 83.86 & 91.22 & 92.65 & 89.84 & 86.96 & 93.03 & 94.82 & 91.30 \\
ChangeCLIP$^*$ & 85.26 & 92.04 & 92.62 & \underline{91.47} & 81.91 & 90.05 & 92.59 & 87.65 \\
ChangeRD & 78.85 & 88.18 & 90.36 & 86.10 & 75.22 & 85.86 & 91.44 & 80.92  \\
CDMamba & 82.38 & 90.34 & 91.57 & 89.15 & 83.71 & 91.13 & 94.93 & 87.63 \\
\hline
\textbf{Ours} & \textbf{85.72} & \textbf{92.31} & 92.47 & \textbf{92.15} & \textbf{89.00} & \textbf{94.18} & \textbf{95.69} & \textbf{92.72} \\
\bottomrule
\end{tabular}
\end{adjustbox}
\end{table}

%% file: tables/exp2.tex
\begin{table}[ht]
\centering
\setlength{\tabcolsep}{3.5pt}
\caption{Quantitative comparisons in terms of IoU, F1, Recall, and Precision on S2Looking-CD and SYSU-CD datasets.}
\label{tab:main2}
\begin{adjustbox}{width=0.48\textwidth}
\renewcommand{\arraystretch}{1.25}
\begin{tabular}{l|cccc|cccc}
\toprule
 & \multicolumn{4}{c|}{\textbf{S2Looking-CD}} & \multicolumn{4}{c}{\textbf{SYSU-CD}} \\
\midrule
\midrule
Methods & IoU & F1 & Pre. & Rec. & IoU & F1 & Pre. & Rec. \\
\midrule
\midrule

FC-Diff & 24.61 & 39.50 & \textbf{80.20} & 26.10 & 42.03 & 59.18 & \textbf{90.31} & 44.01 \\
IFNet & 44.76 & 61.84 & 69.06 & 55.98 & 66.02 & 79.53 & 87.30 & 73.04 \\
BIT & 45.62 & 62.65 & 70.26 & 56.53 & 57.88 & 73.32 & 75.15 & 71.58\\
ChangeFormer & 46.69 & 63.65 & 69.51 & 58.71 & 64.29 & 78.26 & 78.17 & 78.36 \\
A2Net & 48.39 & 65.61 & 69.21 & 61.66 & 71.37 & 83.29 & 86.54 & 80.28 \\
CGNet& 46.78 & 63.74 & 70.72 & 58.02 & 66.55 & 79.92 & 86.37 & 74.37 \\
CLAFA & 48.75 & 65.55 & 71.09 & 60.81 & 70.10 & 82.43 & 84.38 & \underline{80.56} \\
BiFA & 45.70 & 62.73 & 65.15 & 60.49 & \underline{71.73} & \underline{83.53} & 87.05 & 80.29 \\
WS-Net++ & \underline{49.54} & \underline{66.26} & 69.50 & \textbf{63.30} & 70.13 & 82.44 & 87.05 & 78.29 \\
ChangeCLIP$^*$ & - & - & - & - & 70.46 & 82.67 & 84.89 & \underline{80.56} \\
ChangeRD & 29.73 & 45.84 & 62.10 & 36.33 & 59.38 & 74.52 & 79.87 & 69.83  \\
CDMamba & 44.85 & 61.92 & 65.44 & 58.77 & 65.72 & 79.32 & 81.01 & 77.69 \\
\hline
\textbf{Ours} & \textbf{50.52} & \textbf{67.13} & \underline{71.63} & \underline{63.17} & \textbf{73.46} & \textbf{84.70} & \underline{87.87} & \textbf{81.75} \\
\bottomrule
\end{tabular}
\end{adjustbox}
\end{table}    

%% file: tables/ablation.tex
\newcolumntype{C}[1]{>{\centering\arraybackslash}m{#1}}

\begin{table}[ht]
\centering
\caption{Ablation studies of the proposed components on the WHU\mbox{-}CD and LEVIR\mbox{-}CD datasets. \ding{51} and \ding{55} denote "w/" and "w/o" the specific module.}
\label{tab:abl}
\resizebox{\linewidth}{!}{
\begin{tabular}{C{11.3mm}C{8mm}C{8mm}|C{10mm}C{10mm}|C{10mm}C{10mm}}
\toprule
\multicolumn{3}{c|}{Components} & \multicolumn{2}{c|}{\textbf{LEVIR-CD}} & \multicolumn{2}{c}{\textbf{SYSU-CD}} \\
\midrule
\midrule
\textbf{DFFM} & $\boldsymbol{\mathrm{S^2DT}}$ & \textbf{LMM} & IoU & F1 & IoU & F1 \\
\midrule
\midrule

\ding{55} & \ding{55} & \ding{55} & 84.23 & 91.44 & 70.06 & 82.40 \\
\ding{55} & \ding{51} & \ding{51} & 84.49 & 91.59 & 71.61 & 83.46 \\
\ding{51} & \ding{55} & \ding{51} & 84.98 & 91.88 & 72.87 & 84.30 \\
\ding{51} & \ding{51} & \ding{55} & 85.65 & 92.27 & 72.54 & 84.09 \\
\midrule
\ding{51} & \ding{51} & \ding{51} & \textbf{85.72} & \textbf{92.31} & \textbf{73.46} & \textbf{84.70} \\
\bottomrule
\end{tabular}
}
\end{table}



%% file: sections/conclusion.tex
\section{Conclusion}\label{CONCLU}

In this work, we proposed \textbf{ChangeDINO}, an end-to-end framework for optical building change detection that combines a Siamese backbone, DINOv3-pretrained multi-scale features, a spatial–spectral differential transformer decoder, and a learnable morphology head. The DINOv3 branch provides semantically strong, domain-agnostic features for small RSCD datasets, the $\mathrm{S}^2\mathrm{DT}$ decoder uses change priors to emphasize true changes and suppress artifacts, and the morphology module refines boundaries. Experiments on four public benchmarks show that ChangeDINO outperforms recent CNN-, Transformer-, and foundation-model–based methods in IoU and F1, and ablations confirm the contribution of each component. In future work, we plan to extend the framework to multi- and hyperspectral remote sensing data and to related tasks such as land-cover change analysis and UAV-based urban monitoring under the deep learning paradigm.